\documentclass[11pt]{article}
\usepackage{acl2015}
\usepackage{times}
\usepackage{url}
\usepackage{latexsym}
\usepackage{graphicx}
\usepackage{array}
\usepackage{lscape}

\usepackage{amsmath}
\usepackage[utf8]{inputenc}
\usepackage[english]{babel}
\usepackage{natbib}
\title{Sentiment Classification using Images and Label Embeddings}

\author{Laura Graesser \\
  {\tt lhg256@nyu.edu} \\\And
  Abhinav Gupta \\
  {\tt abhinavg@nyu.edu} \\\And
  Lakshay Sharma \\
  {\tt ls4170@nyu.edu} \\\And
  Evelina Bakhturina \\
  {\tt eb2992@nyu.edu} \\}  

\begin{document}
\maketitle
\begin{abstract}

 In this project we analysed how much semantic information images carry, and how much value image data can add to sentiment analysis of the text associated with the images. To better understand the contribution from images, we compared models which only made use of image data, models which only made use of text data, and models which combined both data types. We also analysed if this approach could help sentiment classifiers generalize to unknown sentiments.

\end{abstract}

\section{Introduction}

Nowadays, people tend to use images, videos, and GIFs to express their emotions and opinions alongside text. For example, images shared on websites such as Flickr and Instagram are often accompanied with a short description to augment or further explain the post. This hybrid, or multi-modal, data is particularly prevalent on social media, where users frequently express their opinions on a wide range of topics. 

Accurate sentiment classification has many applications. It can help companies understand how customers feel about their products, politicians understand how people are responding to a policy proposal, or be used to investigate cultural differences \citep{redi}. Effectively making use of multi-modal data in sentiment classification, images and text for example, is valuable since it expands the sources of information available to organizations to understand the people who are engaging with them or their products.

Sentiment classification is both an interesting and a hard problem due to the inherent subjectivity in identifying sentiment. Perception of sentiment is often shaped by one's culture. Different cultures exhibit varying degrees of positivity \citep{redi}, and the same description or image may evoke a different sentiment depending on where the person you ask is from. Furthermore, sentiment classification is hard because it can be ambiguous. The same image can evoke mixed feelings in a viewer. It is not a contradiction to look at an image and feel happy and sad at the same time. This notion is also reflected in our language, with phrases such as "bittersweet", which nicely capture the ambiguity of sentiment. Consequently, most datasets that have been gathered for sentiment analysis are either small or only weakly labeled. As a result, they tend to be noisy, with examples exhibiting large variation within sentiment classes, as well as some examples being mislabeled.

Up until now, most sentiment analysis research has been focused on text. Whilst this is still a hard problem, significant progress has been made in this direction (for example \cite{socher2013recursive}). However, the increasing use of multiple modes to express opinions means that it is important to try and learn to recognise sentiment using sources of information other than text. Ultimately, the goal is to combine each of these sources of information to produce a more nuanced and accurate picture of sentiment.

The objective of this project was to determine whether the use of images was helpful in identifying the sentiment of multi-modal data consisting of images and text. Since sentiment classification is often not clear cut, we also looked for a way to better understand the outputs of our models. To do so, we took inspiration from the increased use of word embeddings, a dense representation of a word in vector space which encourages information sharing between similar words, across a wide range of natural language processing tasks, and the application of this idea to output / label space as in \cite{devise} and \cite{socher}. By forcing our models to output their results in an embedded space, we were better able to assess how the predicted sentiment of different examples from our data related to each other, and whether they aligned with our intuition about which datapoints should have been difficult or easy to classify. 

A final goal of this project was to see if this approach could help sentiment classifiers generalize to unknown sentiments. Our final models were not able to generalize to unknown classes. However, we believe this is an interesting area for further work and that models which output results into an embedded space are a step in the right direction.

\section{Dataset}

\subsection{Original Dataset}\label{or_dataset}
Our work was based on the large-scale multilingual visual sentiment concept ontology (MVSO) \citep{MVSO}.  
This dataset consists of more that 7.36 million xml files with Flickr images metadata. The original dataset contains multilingual data (12 languages), for our project we restricted ourselves to the English subset of the data.

Each xml file contains the following information:
\begin{itemize}
    \item image URL
    \item text fields: title, description, tags
\end{itemize}

\begin{figure}[h!]
\includegraphics[scale=.22]{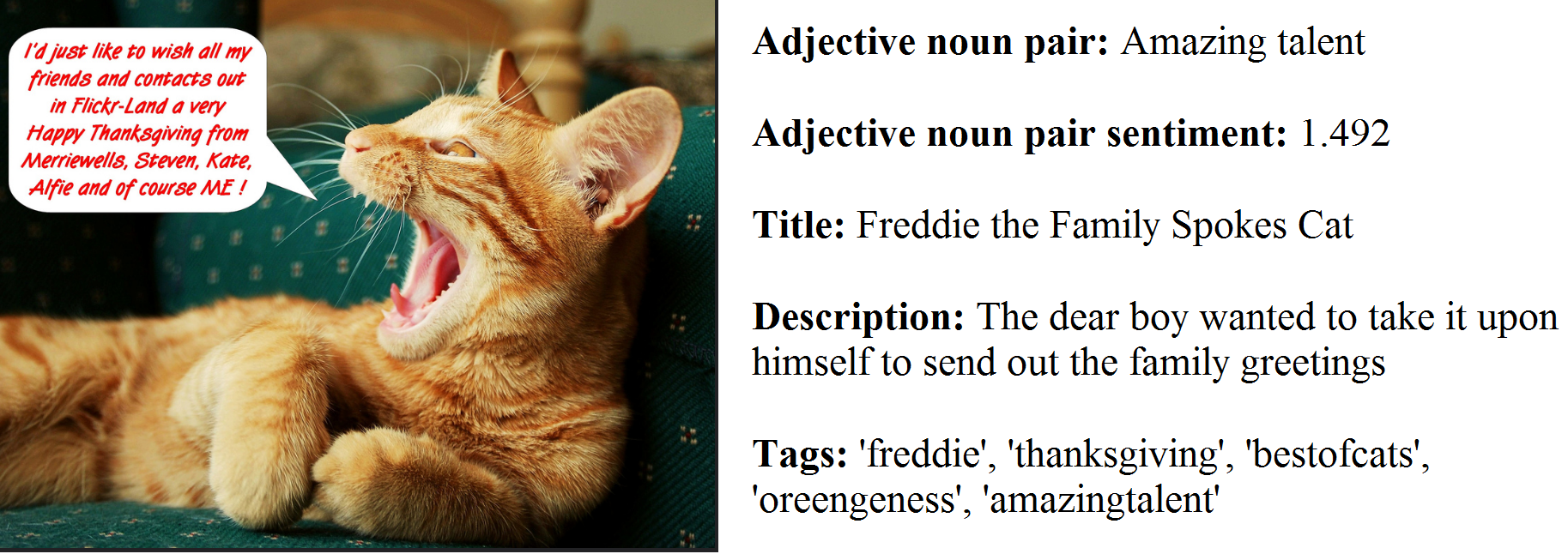}
\caption{Data point example}
\label{fig:data}
\end{figure}

Each xml file is associated with a related sentiment-based adjective-noun with a sentiment score. For example, ANP "infinite sadness" has sentiment score -2.128, while ANP "nice smile" has an associated score 2.019. The English subset of the data contains about 4200 distinct English ANPs. The example of a data entry is shown in Figure \ref{fig:data}.

\subsection{Data Preprocessing}
We applied a number of filters to the original dataset, and kept only files that satisfy the following criteria:

\begin{itemize}
    \item the associated image is still accessible
    \item text fields are in English and spelled correctly
    \item text fields combined contain at least 10 words
\end{itemize}

As a result of the data preprocessing, the final dataset contained about ~186k data points.

For each experimental setup that follows, the data (after further pruning, if any), was shuffled and split into training (70\%), validation (20\%), and test sets (10\%).

\subsection{Evaluation metric}

The primary metric we used to compare models was accuracy.
It worth noting that for the Embedding Classifier (thorough description of Embedding Classifier is given in Section \ref{related}) cosine distance was used to measure similarity between the generated word embeddings and true labels embeddings. 

\subsection{Feature Extraction from images}\label{alexnet}
As an input for some of our models, we used features extracted from images by running our image data though a state-of-art pre-trained models: AlexNet \citep{alexnet} and ResNet \citep{ResNet}. These models were trained for image recogintion/classification task, and since many first layers of the neural networks learn image specific information (edges, shape, colors, etc.) before learning to classify them, we utilized these models' architecture to make our models less complex.

\section{Related Work/Baselines}\label{related}
Only a limited amount of work has been done on sentiment analysis of images. The works that inspired our project were \cite{devise} and \cite{yang}. 
\cite{devise} proposed a model that matches a-state-of-art performance on 1000-class ImageNet object recognition challenge by utilizing semantic information learned from text labels. \cite{devise} used a pre-trained neural language model and a pre-trained state-of-the-art deep neural network for visual object recognition complete with a traditional softmax output layer. Then they took the lower layers of the pre-trained visual object recognition network and re-trained them to predict the vector representation of the image label text as learned by the language model. In this work, the final model was capable of classifying unseen classes by generating word embedding of the ‘predicted’ class text label.

\cite{yang} performed a sentiment classification of Flickr and Twitter images with a Convolutional Neural Networks (CNN) model. To the best of our knowledge \cite{yang} were the first to apply deep learning methods to image sentiment classification, and achieved state of the art results by doing so. 

\cite{luo}, and \cite{you} explore image-text sentiment classification. \cite{you} extracted image features using a convolutional neural network and text features using a doc2vec model. They compare three different approaches for combining models, concatenating both sets of features and building a logistic regression model on top of the features, taking the average of the predicted sentiment score across the two models, and cross-modality consistent regression, with the latter achieving the best results. \cite{luo} take a different approach, using a tree structured LSTM to combine image and text features, and achieved state of the art results for combined textual-visual sentiment analysis on this dataset. 

We used a similar approach to \cite{you} to combine our models, but differ by using an LSTM model to generate the text features. Additionally, we explore the use of gates to control how much information from the text and image features are used for the final classification task.

Finding a well labeled sentiment dataset is challenging. There were no readily available large datasets that contain images with associated captions and the sentiment labels. We therefore used the same dataset as \cite{yang} to conduct our experiments, the MVSO dataset. \cite{yang} used the ANP pairs and associated scores to assign labels to particular datapoints, although they are not specific about how they translated from ANP score to sentiment labels. Furthermore, this approach could be improved upon since datapoints are only labeled indirectly, due to the ANP pair that is associated with it, rather than due to the specific content of the image and text. That is why, our initial idea was to label the dataset on our own (Section \ref{label1}) by labeling the images with the sentiment label the associated caption carries. As we discuss in Section \ref{label1}, the generated labels were not reliable, requiring us to filter our dataset to a much smaller set to try and improve the reliability of our training examples, and the results we obtained did not approach state of the art. Therefore, we decided to try another labeling approach Section \ref{label2} and use ANP pairs and associated sentiment scores to label image-caption pairs. This is the same approach taken by \cite{yang} and \cite{you}.

The most comparable set of experiments to ours were from \cite{luo}. Their methodology for gathering images was very similar, applying filters to exclude examples with no image or text for example. Their dataset was also the most comparable size compared to ours, ~250k datapoints, split almost equally between positive and negative classes. Other research that utilized the same dataset, \cite{yang} for example, used a much larger set of 0.5 million images. We therefore use results of \cite{luo} as benchmarks for our image only, text only, and combined models. We also include \cite{yang}'s results for image sentiment classification as a state of the art comparison for this task. Finally, whilst we experimented with both two (positive, negative), and three (positive, negative, neutral) sentiment classifiers, \cite{yang} and \cite{luo} only provide results for two classes. Consequently, we have no benchmarks for our three class sentiment classifiers.

To summarize, our project compares with related work in the following ways:
\begin{enumerate}
    \item {As in \cite{yang}, we experiment with sentiment image classification (Classical classifier) }
    \item {We incorporated ideas from the \cite{devise} architecture and trained our models to predict the vector representation of the sentiment label (Embedding classifier).}
    \item {As in \cite{luo} paper, we experiment with sentiment image - text classification (Classical and embedding classifier)}
    \item {To the best of our knowledge, no one has yet combined sentiment analysis of images, or images and text, with output to an embedding space.}
\end{enumerate}

\section{Our Approach}

Broadly, our experiments are split up into two parts. The sub-sections that follow are divided as per these two parts.

\begin{enumerate}
\item In the first set of experiments, for classification, we use only image features as input. Within this set, there are two different labeling approaches
\item In the second set of experiments, we also include textual information as input.
\end{enumerate}

The Classical Classifier and Embedding Classifier networks that we created were relatively simple feed-forward networks. The assumption for using these was that, since the image features are obtained from a very sophisticated network like AlexNet, these features would be sufficiently elaborate to provide enough distinguishing power to our classifier networks.

\subsection{Using only image features as input}

For this section, only features obtained from the image are used as input. The features used here are 4096-dimensional, obtained using AlexNet.

\subsubsection{Using textual information from image metadata for labeling}\label{label1}

In this labeling approach, we use existing sentiment prediction tools to examine the image's title and description fields, and generate labels based on these inputs.

The first tool we tried for labeling the dataset was Stanford NLP sentiment predictor from the Stanford Core NLP Package \footnote{http://nlp.stanford.edu/sentiment/}. This tool works by building a parse tree for each given sentence, and using the parse tree to generate sentiment scores/labels for the sentences (\cite{stanfordnlp} for details).

The Stanford package assumes that the incoming sentences have proper punctuation and sentence structuring, while in our dataset even after the pre-processing most of the text fields represent short phrases that lack proper punctuation and structure. Analysis of the produced labels made us disregard labels generated by this package and try Google Cloud Natural Language API \footnote{https://cloud.google.com/natural-language/} instead.



This tool provides sentiment scores for given text. We manually set thresholds in the range of possible scores to distinguish and assign sentiment labels. It was easier to generate fairly accurate labels for each multi-sentence text input using this tool, as compared to the previous one. This also proved to be much faster at generating labels.

However, as is the case with any machine labeling scheme, the labeled data contained quite a lot of noisy data. This also produced an overwhelmingly very large number of neutral labels (about 70\%). In order to overcome this and have approximately equally balanced classes, the dataset was reduced to about 15.2K samples.

\textbf{Softmax classifier network (Classical Classifier)}\label{softmax1}
\\
The first type of network we use is a feed-forward neural network, which takes as input a 4096-dimensional vector of image features obtained from the prerained AlexNet (section \ref{alexnet}), and output a softmax layer, with every node representing a probability of a particular sentiment class \footnote{For all the experiments we conducted we used Keras \cite{keras}}.

We tried a variety of hidden layer architectures, and our experimental results are summarized in the sub-sections below.

Fixed parameters (unless otherwise specified):
\begin{itemize}
    \item Data-points: approximately 5K per class 
    \item number of epochs = 20
    \item batch size = 64 
    \item loss function = categorical cross-entropy
    \item optimizer = stochastic gradient descent
    \item learning rate = 0.001 
    \item momentum = 0.9
\end{itemize}

Experiments with \textbf{3 classes (positive, neutral, negative)} are shown in Table \ref{3cl}.
\begin{table}[h]
\small
\begin{center}
\begin{tabular}{ | m{14.8em} | m{1.6cm}| }
\hline \bf Architecture & \bf Validation Accuracy \\\hline
In = 4096 ReLU, Dns = 512 ReLU, D-out = 0.2, Dns = 3 S-max & .400 \\ \hline
In = 4096 ReLU, Dns = 1024 ReLU, D-out = 0.5 Dns = 512 ReLU, D-out = 0.5, Dns = 3 S-max & \textbf{.434} \\\hline
In = 4096, ReLU, Dns = 1024 ReLU, D-out = 0.5 Dns = 512 ReLU, D-out = 0.5, Dns = 3 S-max & .422 \\\hline
In = 4096, Dns = 512 ReLU, D-out = 0.2, Dns = 3 S-max & .424 \\\hline
\end{tabular}
\end{center}
\caption{Classical Classifier results with 3 classes\\
In = Dimension of Input, Dns = Dense layer, D-out = dropout, S-max = Softmax, ReLU - Rectifier}
\label{3cl}
\end{table}

Experiments with \textbf{2 classes (positive, negative)} are shown in Table \ref{2cll}. 

\begin{table}[h]
\small
\begin{center}
\begin{tabular}{ | m{14.8em} | m{1.6cm}| }
\hline \bf Architecture & \bf Validation Accuracy \\\hline
In = 4096, ReLU, Dns = 2048 ReLU, D-out = 0.7 Dns = 1024 ReLU D-out = 0.7 Dns = 512 ReLU, D-out =0.5, Dns = 3 S-max
(30 epochs) – Model 1 & .616 \\ \hline
In = 4096, ReLU, Dns = 2048 ReLU, D-out = 0.7 Dns = 1024 ReLU D-out = 0.7 Dns = 512 ReLU, D-out =0.5, Dns = 3 S-max
(60 epochs) – Model 2 & \textbf{.633} \\\hline
Fine tuned Model 1 with samples having prob $>$ 0.6
(20 epochs) & .617 \\\hline
Fine tuned Model 2 with samples having prob $>$ 0.6
(30 epochs) & \textbf{.633} \\\hline
\end{tabular}
\end{center}
\caption{Classical Classifier results with 2 classes\\
In = Dimension of Input, Dns = Dense layer, D-out = dropout, S-max = Softmax, ReLU - Rectifier}
\label{2cll}
\end{table}

In both, the two class and three class classification schemes, it was observed that fitting the training data was not difficult, with training accuracy almost always being above 95\%, and often close to 100\%. However, as the validation results show, the model overfits and fails to generalize. Attempts to fight overfitting with regularization, dropout in particular, did not prove to be useful. The reason for this could be the relatively small training data set we got after balancing the classes. The softmax classifier on 2 sentiment classes had an accuracy of .633. While not as accurate as we had hoped, this classifier was able to achieve accuracy better than the visual classifier implemented by \cite{luo}, which also used the MVSO dataset, and achieved an accuracy of .616.

This confirms our initial hypothesis that relatively simple feed-forward networks have enough capacity, given sufficiently good features, to fit a fairly complex training set, but generalizing is still a challenge.

\textbf{Embedding Classifier network}\label{embedding1}
\\
Besides the Classical Classifier, which is trained to predict a set of classes, we trained a model to predict a dense (vector) representation of sentiment label. The word embeddings for each label (positive, neutral, negative) were obtained from the set of pre-trained word vector representations provided by GloVe: Global Vectors for Word Representation  \citep{glove}. We used word embeddings of dimension 50 obtained from the model trained on Wikipedia 2014 + Gigaword 5 corpus\footnote{http://nlp.stanford.edu/projects/glove/}.
In the embedding classifier we also used a feed-forward neural network, which  takes a 4096-dimensional vector of image features generated with the AlexNet as input and outputs a 50-dimensional projection vector associated with that particular label (by this, we mean the textual/word representation of the label). At validation/testing time, the label assigned is the one whose embedding is closest to this 50-dimensional output.

We tried a variety of hidden layer architectures, and our experimental results are summarized in the sub-sections below.
\\
\\
Fixed parameters:
\begin{itemize}
\item Data-points: approximately 5K per class
\item number of epochs = 20
\item batch size = 64
\item loss function = cosine proximity
\item optimizer = root mean square propagation
\item embedding similarity measure = cosine similarity
\end{itemize}

Experimental results with \textbf{3 classes (positive, neutral, negative)} are shown in Table \ref{3cl2}.
\begin{table}[h]
\begin{center}
\begin{tabular}{ | m{14.8em} | m{1.6cm}| }
\hline \bf Architecture & \bf Validation Accuracy \\\hline
In = 4096, ReLU, Dns = 512 ReLU, D-out =0.2, Dns = 50 linear & .412 \\ \hline
In = 4096, ReLU, Dns = 512 ReLU, D-out =0.2, Dns = 50 linear & \textbf{.430} \\\hline
In = 4096, ReLU, Dns = 1024 ReLU, D-out = 0.5 Dns = 512 ReLU, D-out =0.5, Dns = 50 linear & .424 \\\hline
\end{tabular}
\end{center}
\caption{Embeddding Classifier results with 3 classes\\
In = Dimension of Input, Dns = Dense layer, D-out = dropout, S-max = Softmax, ReLU - Rectifier\\}
\label{3cl2}
\end{table}

Experimental results with \textbf{2 classes (positive, negative)} are shown in Table \ref{2cl2}.
\begin{table}[h]
\small
\begin{center}
\begin{tabular}{ | m{14.8em} | m{1.6cm}| }
\hline \bf Architecture & \bf Validation Accuracy \\\hline
In = 4096, ReLU, Dns = 2048 ReLU, D-out = 0.7 Dns = 1024 ReLU D-out = 0.7 Dns = 512 ReLU, D-out =0.5, Dns = 50 S-max
(30 epochs) & \textbf{.617} \\ \hline
In = 4096, ReLU, Dns = 2048 ReLU, D-out = 0.7 Dns = 1024 ReLU D-out = 0.7 Dns = 512 ReLU, D-out =0.5, Dns = 50 S-max
(60 epochs) & .613 \\\hline
\end{tabular}
\end{center}
\caption{Embedding Classifier results with 2 classes\\
In = Dimension of Input, Dns = Dense layer, D-out = dropout, S-max = Softmax, ReLU - Rectifier\\}
\label{2cl2}
\end{table}

The results with the embedding-based classifier were similar to the softmax classifier (classical classifier): models overfitted the training set and could not generalize the validation data. This classifier performed slightly better (accuracy: .617) than the visual classifier implemented by \cite{luo}, also using the MVSO dataset (accuracy: .616).

We hypothesized that the reasons for the models not being able to generalize could be: a result of noise in the machine-labeled data, and small size of the training dataset. To remedy this, our next approach was to try a new labeling technique, which also yields a larger training dataset. 
 
\subsubsection{Using ANP scores to label the dataset}\label{label2}
To combat noise in the machine-labeled data, we tried labeling using adjective-noun pairs. This appears to be a popular labeling technique, used in several related works.

As was mentioned in Section \ref{or_dataset} that every image in the MSVO dataset is associated with a adjective-noun pair (ANP) and a corresponding sentiment score. 

The sentiment scores fall in a range from -2.022 to 2.16. To obtain well-balanced classes and get well defined labels (high sentiment score is associated with highly positive label, and vice versa highly negative sentiment score is associated with the negative label; the neutral labels are associated with scores in the zero region) the following way:
\begin{itemize}
\item positive class has sentiment score from 2.16 to 0.035
\item neutral class has sentiment score from 0.034 to -0.034
\item negative class has sentiment score -2.022 to -0.035
\end{itemize}

These thresholds resulted in the equally-balanced classes. 

Note: in this approach, in addition to having as input 4096-dimensional feature vectors obtained from AlexNet, we have also, in some places, used raw pixel data of dimensions 32*32*3 from the images. The input type is specified when describing the results.

\textbf{Softmax classifier network}
\\
As was the case in \ref{softmax1}, the first type of network we use is a feed-forward neural network, which takes as input a 4096-dimensional vector of image features, and has as output a softmax layer, with every node representing a sentiment class.

We tried a variety of hidden layer architectures, and our experimental results are summarized in the sub-sections below.

Fixed parameters (unless otherwise specified):
\begin{itemize}
\item Data-points: approximately 38K per class
\item epochs = 40
\item batch size = 64
\item loss function = categorical cross-entropy
\item optimizer = stochastic gradient descent
\item learning rate = 0.001
\item momentum = 0.9
\end{itemize}

Experimental results with \textbf{3 classes (positive, neutral, negative)} are shown in  Table \ref{4cl3}.

\begin{table}[h]
\begin{center}
\begin{tabular}{ | m{14.8em} | m{1.6cm}| }
\hline \bf Architecture & \bf Validation Accuracy \\\hline
In = 4096, Dns = 2048 ReLU, Dns = 1024 ReLU, Dns = 512 ReLU, Dns = 24 ReLU, Dns = 3 S-max & .420 \\ \hline
In = 4096, Dns = 2048 ReLU dropout=0.5, Dns = 1024 ReLU dropout=0.5, Dns = 512 ReLU dropout=0.5, Dns = 24 ReLU = dropout=0.5, Dns = 3 S-max & .402 \\\hline
In = 4096, Dns = 2048 ReLU dropout=0.2, Dns = 1024 ReLU dropout=0.2, Dns = 512 ReLU dropout=0.2, Dns = 24 ReLU, Dns = 3 S-max & \textbf{.426} \\\hline
\end{tabular}
\end{center}
\caption{Classical Classifier results with 3 classes\\
In = Dimension of Input, Dns = Dense layer, D-out = dropout, S-max = Softmax, ReLU - Rectifier}
\label{4cl3}
\end{table}

Experimental results with \textbf{2 classes (positive, negative)} are shown in Table \ref{4cl2}.
\begin{table}[h]
\small
\begin{center}
\begin{tabular}{ | m{14.8em} | m{1.6cm}| }
\hline \bf Architecture & \bf Validation Accuracy \\\hline
In = 32* 32 * 3, Conv 96, MaxPool, BatchNorm, Conv 256, MaxPool, BatchNorm, Dns = 1024 ReLU, Dns = 512 ReLU, Dns = 24 ReLU, Dns = 2 S-max & .545 \\ \hline
In = 32* 32 * 3, Convolution 96, MaxPool, BatchNorm, Conv 256, MaxPool, BatchNorm, Dns = 1024 ReLU, D-out  = 0.5, Dns = 512 ReLU, D-out  = 0.5, Dns = 24 ReLU, D-out  = 0.5, Dns = 2 S-max & .506 \\\hline
In = 4096, Dns = 2048 ReLU, Dns = 1024 ReLU, Dns = 512 ReLU, Dns = 24 ReLU, Dns = 2 S-max & .595 \\\hline
In = 4096, Dns = 2048 ReLU, D-out  = 0.5, Dns = 1024 ReLU, D-out  = 0.5, Dns = 512 ReLU, D-out  = 0.5, Dns = 24 ReLU, D-out  = 0.5, Dns = 2 S-max & \textbf{.598} \\\hline
\end{tabular}
\end{center}
\caption{Classical Classifier results with 2 classes
In = Dimension of Input, Conv = Convolution layer, Dns = Dense layer, D-out = dropout, S-max = Softmax, ReLU - Rectifier}
\label{4cl2}
\end{table}

\textbf{Embedding Classifier}
\\
Similar to the \ref{embedding1}, the second type of network we try is also a feed-forward neural network, which  takes as input a 4096-dimensional vector of image features. The output is a projection layer of 50 nodes. 


We tried a variety of hidden layer architectures, and our experimental results are summarized in the sub-sections below.

Fixed parameters (unless otherwise specified):
\begin{itemize}
\item approximately 38K per class
\item batch size = 64
\item number of epochs = 10
\item optimizer = root mean square propagation
\item embedding similarity measure = cosine similarity 
\end{itemize}

Experimental results with \textbf{2 classes (positive, negative)} are shown in Table \ref{5cl2}.
\begin{table}[h]
\begin{center}
\begin{tabular}{ | m{14.8em} | m{1.6cm}| }
\hline \bf Architecture & \bf Validation Accuracy \\\hline
In = 4096, ReLU, Dns = 2096 ReLU, D-out = 0.5, Dns = 1048 ReLU dropout 0.5, Dns = 512, ReLU, D-out = 0.5, Dns = 24 ReLU D-out = 0.2, Dns = 50, linear
(loss = cosine proximity) & \textbf{.618} \\ \hline
In = 4096, ReLU, Dns = 2096 ReLU, D-out = 0.5, Dns = 1048 ReLU dropout 0.5, Dns = 512, ReLU, D-out = 0.5, Dns = 24 ReLU D-out = 0.2, Dns = 50, linear
(loss = mean squared error) & .490 \\\hline
In = 4096, ReLU, Dns = 2096 ReLU, D-out = 0.5, Dns = 1048 ReLU dropout 0.5, Dns = 24 ReLU D-out = 0.2, Dns = 50, linear
(loss = cosine proximity) & .612 \\\hline
In = 4096, ReLU, Dns = 2096 ReLU, D-out = 0.5, Dns = 1048 ReLU dropout 0.5, Dns = 24 ReLU D-out = 0.2, Dns = 50, linear
(loss = mean squared error) & .490 \\\hline
\end{tabular}
\end{center}
\caption{Embedding Classifier results with 2 classes}
In = Dimension of Input, Dns = Dense layer, D-out = dropout, S-max = Softmax, ReLU - Rectifier\\
\label{5cl2}
\end{table}

\textbf{Summary of ANP labeling approach with Classical and Embedding classifiers}

Image features extracted from AlexNet yielded better results compared to using the raw images. This is consistent with our expectations given how well AlexNet image features have performed on a range of image classification tasks. The cosine proximity loss function was also significantly more effective than mean squared error whilst changes in other features had more minor effects. Furthermore, the results obtained via this approach, which examined adjective-noun pairs associated with each image in order to assign labels, were no better than the previous machine-labeling approach. Once again, fitting to the training data was not difficult, but achieving high accuracy on the validation data was. 

At this point, we moved to combining visual information with textual information.

\begin{table}
\begin{center}
\begin{tabular}{ | m{12.5em} | m{1.6cm}| }
\hline \bf Architecture & \bf Validation Accuracy \\\hline
Best image-only classifier: In = 4096, ReLU, Dns = 2048 ReLU, D-out = 0.7 Dns = 1024 ReLU D-out = 0.7 Dns = 512 ReLU, D-out =0.5, Dns = 3 S-max
(60 epochs)  & \textbf{.633} \\ \hline
Best text-only classifier: In = text sequences, post padding, max len = 101, embedding layer = 200, Bidirectional LSTM layer = 300, Dense = 2  & \textbf{.796} \\ \hline
Feature concatenation + linear layer & .807\\ \hline
TxtGL = GL1, ImgGL = GL1, ImgShape = retained, D-out = none  & .805 \\ \hline
TxtGL = GL1, ImgGL = GL1, ImgShape = retained, D-out = 0.3 & .806 \\\hline
TxtGL = GL1, ImgGL = GL1, ImgShape = retained, D-out = none & .804 \\\hline
TxtGL = GL2, ImgGL = GL2, ImgShape = retained, D-out = none & .806 \\\hline
TxtGL = GL2, ImgGL = GL2, ImgShape = retained, D-out = 0.3 & \textbf{.808} \\\hline
TxtGL = GL1, ImgGL = GL2, ImgShape = compressed, D-out = none & .804 \\\hline
TxtGL = GL2, ImgGL = GL2, ImgShape = compressed, D-out = none & .805 \\\hline
TxtGL = GL2, ImgGL = GL2, ImgShape = compressed, D-out = 0.3 & .804 \\\hline
\end{tabular}
\end{center}
\caption{Combined Test: text + images, 2 classes
\\In = Dimension of Input, Dns = Dense layer, D-out = dropout, S-max = Softmax, TxtGL = gated layer of text model, ImgGl = gated layer of image model, ImgShape = shape of image (compressed, or original)}
\label{combinedtable}
\end{table}

\subsection{Combined model: Images and Text}

To explore the contribution of images in a multi-modal sentiment classifier, we combined image features extracted from a pretrained ResNet neural network  with an LSTM model which used the title and description associated with an image to predict the sentiment. 

We used the 50 layer model for the ResNet and extracted features from the penultimate layer of this network. The features were of dimension 2048. We preprocessed the images since the model takes an input image size of (224x224). So all the images were converted from (32x32) to the required dimension.

To generate our textual model, we experimented with a number of different LSTM architectures and hyperparamters. All text data was preprocessed in the same way. First the title and description were concatenated with a period separating the two components. Then we tokenized the resulting text, retaining punctuation but removing whitespace. Datapoints with more than 101 tokens were truncated to reduce computation time and encourage learning by the model, since it reduces the maximum history that could be stored in the LSTM hidden state. Practically, only a few datapoints have more than 100 tokens, (approximately 700), however some examples were very long. Datapoints with fewer than 101 tokens were postpadded with zeros. Finally we sorted all words based on their frequency and converted each token to an integer, using an adaptation of Yasumasa Miyamoto's (a fellow student in the StatNLP class) code for creating a word dictionary from text files. Final vocabulary size is 103k words. External and internal links occurred quite often and were given special 'href' and 'rel' tokens. Each of the LSTM models consisted of an embedding layer, which learnt a dense representation of dimension 200 per token. All models were trained on a single AWS EC2 c4.4xlarge instance, and each model took 4 - 6 hours to train, with some experiments running in parallel.

We varied the number of layers, directionality of layers (single and bi-directional), size of the hidden state, number of training epochs, and optimizer. We found that increasing the size of the hidden layer and adding bi-directionality improved results, whereas adding more layers and adding dropout did not help. This suggests that the model had sufficient capacity and that any further improvements were likely to come from other sources such as more data or a better treatment of the input data. Our best classical classifier model had a single bidirectional layer with a hidden state size of 300, then a dense layer with a softmax activation function so that the model outputted a probability distribution over the classes. It achieved $79.6\%$ on the validation dataset. 

Since the LSTM models took a long time to train, instead of training the embedding model from scratch, we used our best classical model as the starting point, removed the softmax layer and added a linear projection layer to the 50 dimensional embedding space. All models were trained using stochastic gradient descent with a learning rate of 0.01. We found that a hinge loss trained for 10 epochs yielded the best results, and that adding complexity to the model in the form of additional dense layers before the projection layer significantly degraded performance. Our best result achieved $80.8\%$ on the validation set. Interestingly, projecting the results to embedded space added 1ppt to the model's performance. Furthermore $80.8\%$ significantly outperforms the textual baseline of $62.4\%$ and even outperforms the simplest tree-LSTM combined visual-textual model from \cite{luo}, which achieved $80.4\%$.

To combine both types of data, we extracted text features from our text model by using the output from the layer below the softmax layer. First we concatenated the text and image features and implemented a simple linear regression on top of the concatenated feature vector. This achieved an accuracy of $80.7\%$ on the validation set compared to our best previous classical model result of $79.6\%$. Since the textual model was the best model, we wanted to find an architecture that would make use of this model mostly, but would use information from the image model when it was relevant. We tried to implement this by writing a custom gating layer in Keras that had trainable weights. This layer simply multiplied the inputs with the weights elementwise, producing an output with the same dimensions as the input. We hoped that this would enable the network to learn what information to let through from the image and text features. 

Practically we experimented with two types of gating layer, the first (GL1), had no constraints on the weight values. The second (GL2) constrained the weights to take on values between 0 and 1. Our best model applied GL1 to the textual features, compressed the visual features to a vector of dimension 600 using a dense layer with a relu activation, then applied GL2 to the visual features. After that, the results were concatenated, and dropout of 0.3 was applied before a final dense layer with a softmax activation function to give us our predicted sentiment class. It achieved $80.8\%$ ($80.95\%$ was our best observed result with this model but we were not able to replicate it) on the validation set, a slight improvement over the linear model. Given time constraints we restricted our experiments to two sentiment classes, positive and negative. A full summary of our results is displayed in the table number 8.

Finally, to project the model to embedding space, we used the output from the layer below the softmax of our best model. Unfortunately, our results were slightly below that of the classical classifier and our best text-only embedding model, with an accuracy of $79.8\%$. However this result still compares favourably with other reported combined models, outperforming the early, late fusion and CCR combined models, which had an accuracy of $62.1\%$, $65.0\%$ and $67.2\%$ respectively. Table 9 contains a summary of our results.

\begin{table}[h]
\begin{center}
\begin{tabular}{ | m{14.8em} | m{1.6cm}| }
\hline \bf Hyperparameters & \bf Validation Accuracy \\\hline
loss=mse, 5 epochs & .784 \\ \hline
loss=cosine, 5 epochs & .793\\ \hline
loss=hinge, 5 epochs & .661\\\hline
loss=mse, 10 epochs & .786 \\\hline
loss=cosine, 10 epochs & \textbf{.798} \\\hline
loss=hinge, 10 epochs & .781 \\\hline
\end{tabular}
\end{center}
\caption{combined textual-visual model, embedding space}
\label{combinedtable}
\end{table}

\section{Best model: Results and visualization}

Our best classical model achieved $80.8\%$ accuracy on the validation set and $80.9\%$ on the test set. It is encouraging that our test set performance is slightly higher than our validation set performance, indicating that we are not overfitting the validation set. Our per class accuracy was $\approx84\%$ for the positive class, and $\approx77\%$ for the negative class. Whilst the accuracy per class is fairly balanced, it shows that our model is slightly better at identifying positive sentiment than negative sentiment. 

Our best embedded model achieved $79.8\%$ accuracy on the validation set and $78.6\%$ on the test set. It also is not overfitting the validation data, but does have more imbalanced per class accuracy on the test set of $\approx89\%$ for the positive class and $\approx68\%$ for the negative class indicating that in the embedded space it is harder to classify negative sentiment.

The best results along with the associated baselines are shown in Appendix 1 Figure \ref{final_comp}.

To further investigate our results, we applied t-distributed stochastic neighbor embedding (t-SNE) to reduce dimensionality of our data. Appendix 2 Figure \ref{lstm_folders} shows the result of our best Embedding classifier trained on 2 classes. The plot depicts, for each data point, the title of the folder in which the data point is contained (images with the same adjective-noun pair are grouped into the same folder), colored based on the true labels (green for positive, black for negative). The model was trained only on the text description of images and, interestingly, was able to cluster data points coming from the same folder. This aligns with our intuition that different folders should share similar, more nuanced, sentiments. It is particularly exciting that the model was able to recognize some of these clusters without having any explicit information about them, and only being trained to output either the positive or negative word embedding. Some folder names are shown in the bold font meaning that data points from the same folder were plotted were close to each other in the vector space. For example, we can see cluster of the following folders: "ecological disaster", "global protest", "wonderful nature", "beautiful day".

\section{Conclusion}
The MVSO dataset is clearly a challenging dataset for image sentiment classification. The chart in Appendix 1 compares our model performance with benchmarks for the same dataset. Whilst we were disappointed not to approach closer to state of the art performance in image only classification, we achieved reasonable results of $63.3\%$ and beat the the baseline accuracy of $61.6\%$ despite a smaller dataset. Our textual model yielded excellent results, and clearly beat any benchmark we could find for sentiment classification using text data for this dataset. Finally, our best combined model accuracy of $80.9\%$ is approaching state of the art accuracy of $83.3\%$ with a much simpler model architecture. However only a small proportion of the contribution appears to be coming from the image component of the model, at most $1.3$ppts. This is one area for further work. Given more time it would have been interesting to further explore under what circumstances the images contain more useful information than text data for the purposes of sentiment classification and how best to take advantage of this. Given more time we would have liked to use the output from our best image model instead of the ResNet features to generate our image features. Following this, a natural next step would be to a use a larger dataset and to use an image model which achieves closer to state of the art results as part of the combined model architecture. 

Plotting our datapoints in sentiment embedding space also yielded some interesting and intuitive results. Based on this it does seem that distances in sentiment embedding space have meaning, and that given more time and data it may be possible to classify an unknown sentiment class using this approach.





\medskip
\bibliographystyle{acl_natbib}
\bibliography{ref}
\onecolumn
\section*{Appendix}
\bigskip
\begin{center}
\chapter{Appendix 1: Model Comparison}\label{final_comp} \\
\bigskip
\includegraphics[scale=0.43]{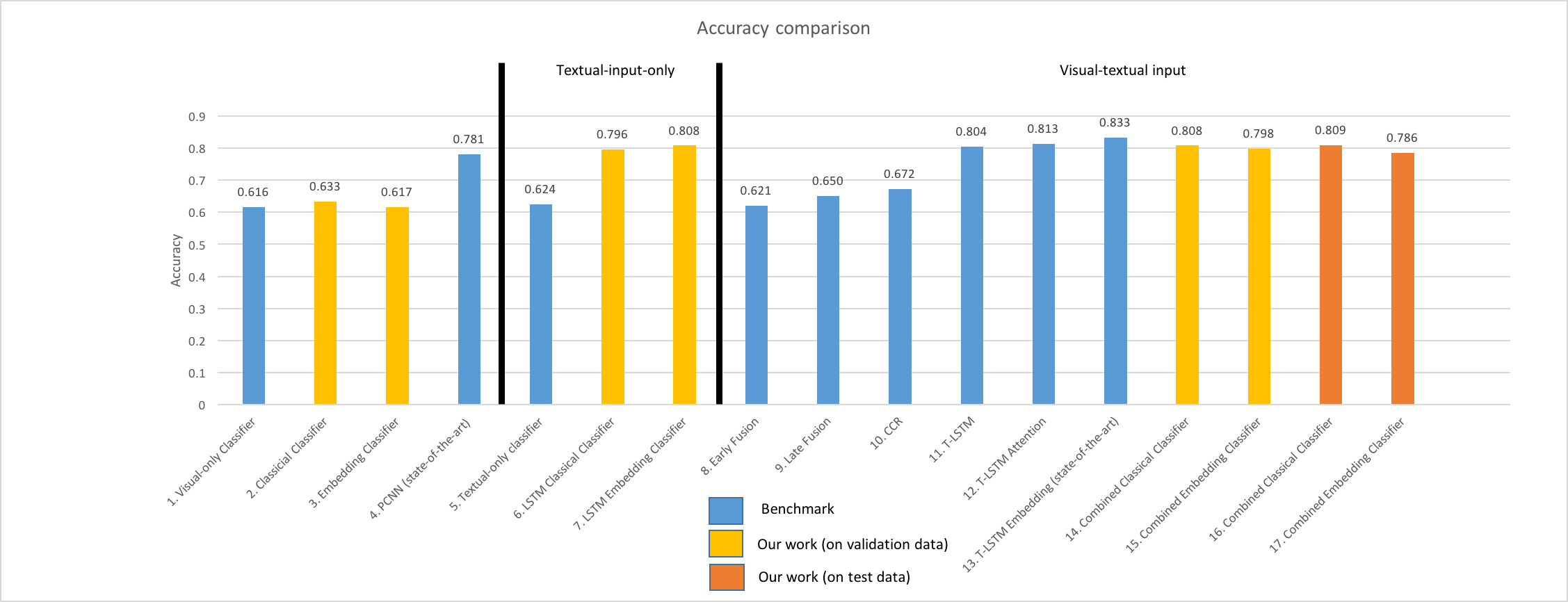}

\begin{enumerate}
\item You et. al. (2016a)
\item Section 4.1.1 Classical Classifier
\item Section 4.1.1 Embedding Classifier
\item You et al. (2015)
\item You et al. (2016a)
\item Section 4.2.1
\item Section 4.2.1
\item You et. al. (2016a)
\item You et. al. (2016a)
\item You et. al. (2016a)
\item You et. al. (2016a)
\item You et. al. (2016a)
\item You et. al. (2016a)
\item Section 4.2.1
\item Section 4.2.1
\item Section 4.2.1
\item Section 4.2.1
\end{enumerate}
\bigskip
\begin{center}
Model Comparison for 2 classes: positive and negative
\end{center}
\label{final_comp}

\newpage
\chapter{Appendix 2: Embedding Classifier Output in Sentiment label embedding space}\label{lstm_foldrs} \\
\bigskip
\centering
\includegraphics[scale=0.42]{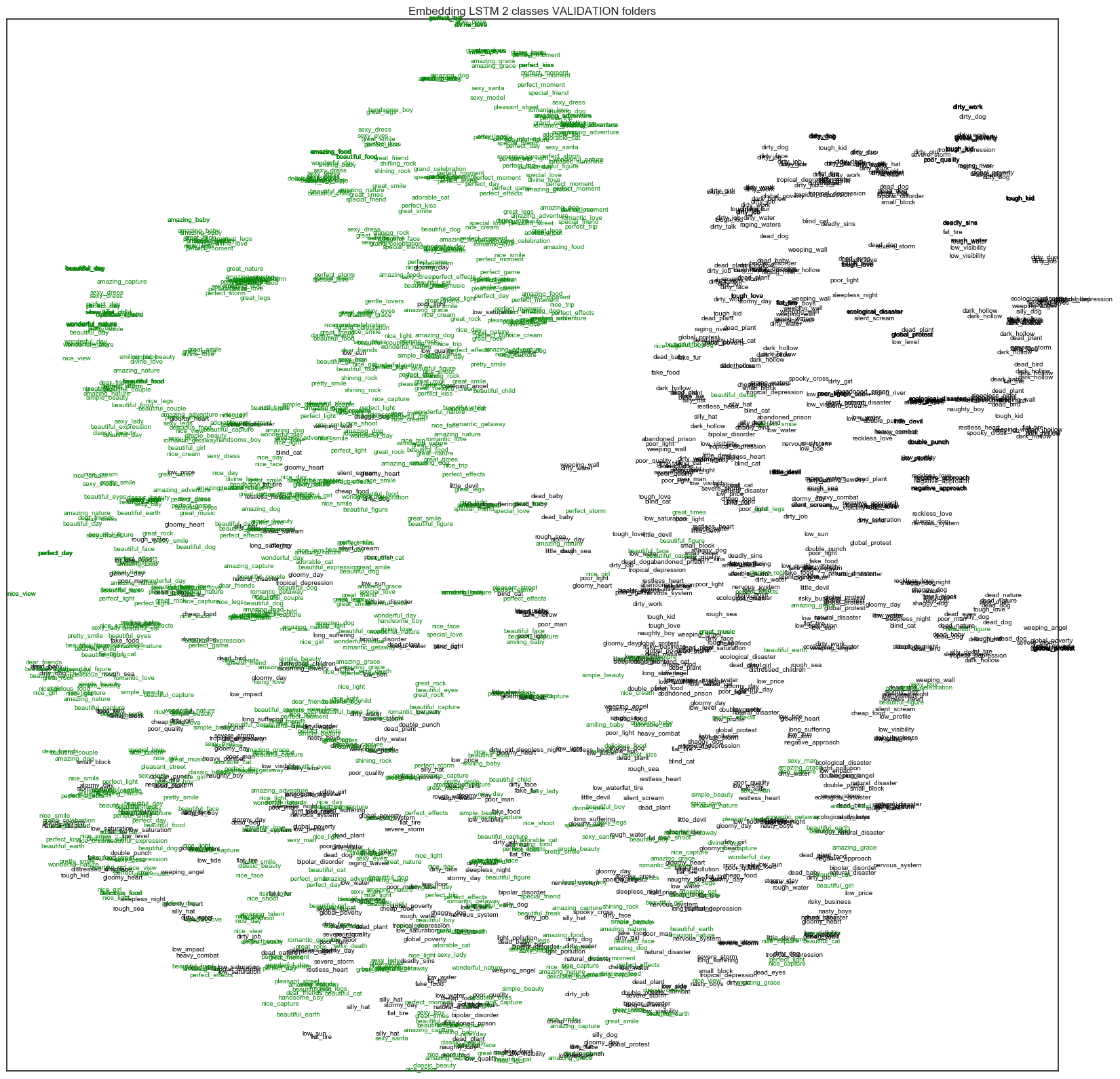}
\label{lstm_folders}

\end{center}
\end{document}